\documentclass{ecai} 
\usepackage{latexsym}
\usepackage{amssymb}
\usepackage{amsmath}
\usepackage{amsthm}
\usepackage{booktabs}
\usepackage{enumitem}
\usepackage{graphicx}
\usepackage{color}
\usepackage{xspace}
\usepackage{multirow}

\newcommand{\isee}{iSee\xspace}

\newcommand{\BibTeX}{B\kern-.05em{\sc i\kern-.025em b}\kern-.08em\TeX}

\begin{document}

\begin{frontmatter}

\paperid{123} 
\title{iSee: Advancing Multi-Shot Explainable AI Using Case-based Recommendations}

\author[A]{\fnms{Anjana}~\snm{Wijekoon}}
\author[A]{\fnms{Nirmalie}~\snm{Wiratunga}\thanks{Corresponding Author. Email: n.wiratunga@rgu.ac.uk}}
\author[A]{\fnms{David}~\snm{Corsar}}
\author[A]{\fnms{Kyle}~\snm{Martin}}
\author[A]{\fnms{Ikechukwu}~\snm{Nkisi-Orji}}
\author[A]{\fnms{Chamath}~\snm{Palihawadana}}
\author[B]{\fnms{Marta}~\snm{Caro-Martínez}}
\author[B]{\fnms{Belen}~\snm{Díaz-Agudo}}
\author[C]{\fnms{Derek}~\snm{Bridge}}
\author[D]{\fnms{Anne}~\snm{Liret}}
\address[A]{Robert Gordon University, Aberdeen, Scotland}
\address[B]{Universidad Complutense de Madrid, Spain}
\address[C]{University College Cork, Ireland}
\address[D]{British Telecommunications, France}

\begin{abstract}
Explainable AI~(XAI) can greatly enhance user trust and satisfaction in AI-assisted decision-making processes. Recent findings suggest that a single explainer may not meet the diverse needs of multiple users in an AI system; indeed, even individual users may require multiple explanations. This highlights the necessity for a ``multi-shot'' approach, employing a combination of explainers to form what we introduce as an ``explanation strategy''. Tailored to a specific user or a user group, an ``explanation experience'' describes interactions with personalised strategies designed to enhance their AI decision-making processes. 
The \isee platform is designed for the intelligent sharing and reuse of explanation experiences, using Case-based Reasoning to advance best practices in XAI. The platform provides tools that enable AI system designers, i.e. design users, to design and iteratively revise the most suitable explanation strategy for their AI system to satisfy end-user needs. All knowledge generated within the \isee platform is formalised by the \isee ontology for interoperability. 
We use a summative mixed methods study protocol to 
evaluate the usability and utility of the \isee platform 
with six design users across varying levels of AI and XAI expertise.
Our findings confirm that the \isee platform effectively generalises across applications and its potential to promote the adoption of XAI best practices. 

\end{abstract}

\end{frontmatter}

\section{Introduction}

XAI systems must be able to address a range of explanation needs~(such as transparency, scrutability, and trust) and must do so in a manner that is relevant to a range of stakeholders. It is also now a regulatory requirement in many parts of the world such as the right to obtain an explanation in the EU~\cite{cath2018artificial,gdpr}.
It is essential for an AI system looking to implement XAI to learn from successful past experiences of XAI adaptations that reveal best practices. Case-based Reasoning~(CBR) caters to this need whereby it learns from past experiences~\cite{aamodt1994case,hammond}. 
The \isee platform has proposed utilising the CBR paradigm to capture knowledge and experience from successful adaptations of explainability within AI systems~\cite{wijekoon2023cbr}.

It is increasingly recognised that a single explanation is often insufficient to satisfy all situations and/or all stakeholders~\cite{arya2019one,mohseni2021multidisciplinary}. Multi-shot explanations, allowing users to digest explanations from multiple algorithms over the course of a single interaction, have been demonstrated to provide more satisfactory user experiences~\cite{malandri2023convxai,wijekoon2023cbr}. However, bespoke development of domain-specific and personalised explanation strategies is prohibitively expensive and requires significant domain and XAI expertise. Explainability toolkits have emerged to facilitate the development of explainability for use cases, including IBM-XAI 360~\cite{aix360}, Alibi~\cite{alibi} and Captum~\cite{kokhlikyan2020captum}. Where existing toolkits fall short, \isee fills this gap by providing support to design users, regardless of their level of expertise in XAI, to develop explanation strategies based on the collective experiences of others.

\begin{figure}[t]
\centering
\includegraphics[width=.48\textwidth]{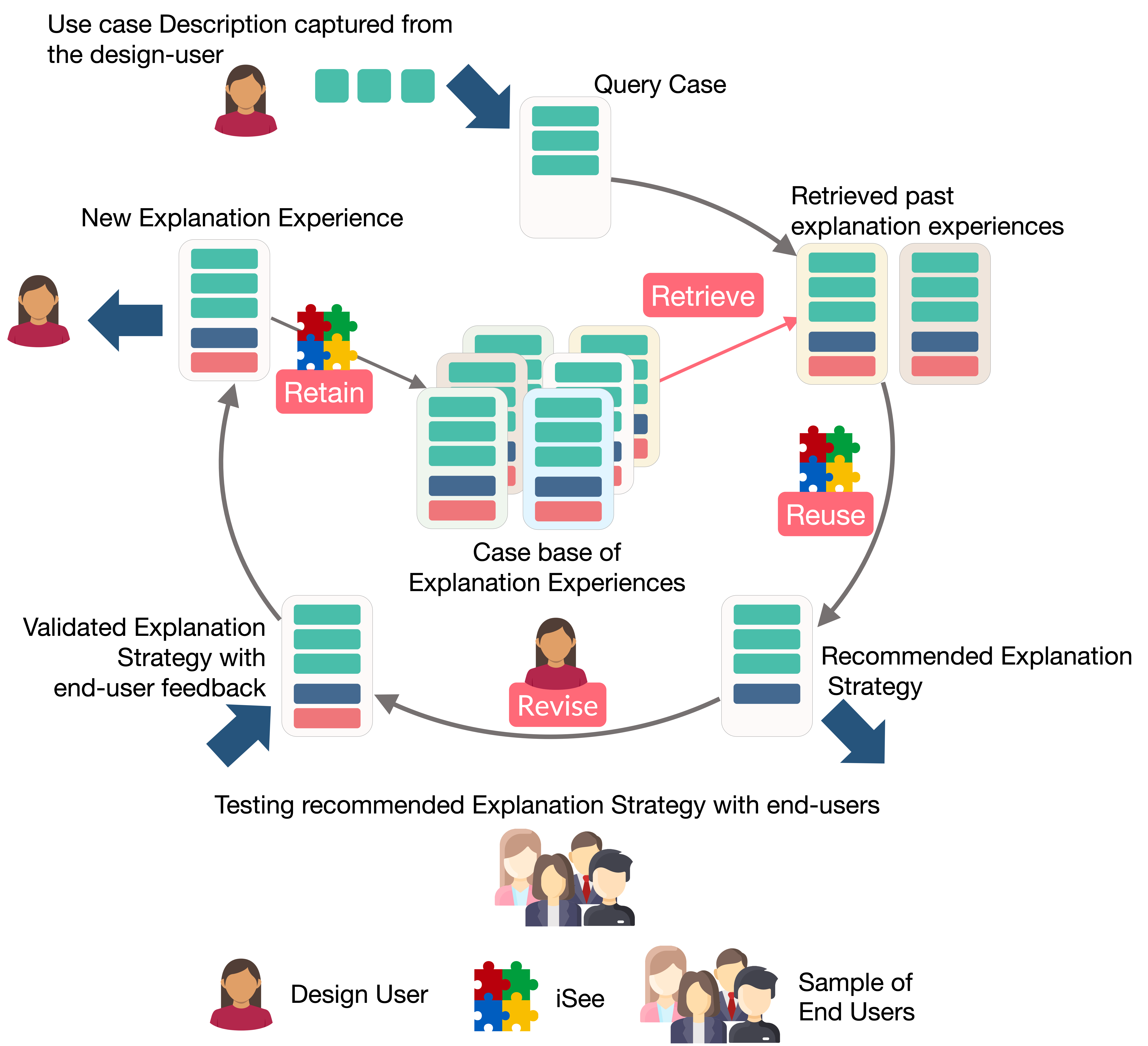}
\caption{\isee CBR Cycle}
\label{fig:cbr}
\end{figure}

The \isee Cockpit is designed to elicit stakeholder requirements for explainability from the design user, which drives the underpinning CBR paradigm of the platform. Accordingly, in this paper, we evaluate the \isee Cockpit tools for capturing these requirements from real-world design users towards the design of multi-shot explanation strategies. 
We make two key contributions: 1) A formalisation of the multi-shot explanation experience underpinned by the CBR paradigm; and
2) The design and findings of a user experience evaluation of the \isee cockpit with insights from six design users, highlighting the tool's usability, utility, and areas for improvement.

The rest of this paper is organised as follows. Related literature is discussed in Section~\ref{sec:related}. Section~\ref{sec:app} presents the \isee platform, and contextualises the system using a radiography fracture detection AI system. Section~\ref{sec:design} presents the user experience study design and results are presented in Section~\ref{sec:results}, followed by conclusions in Section~\ref{sec:conc}. 

\section{Related Work}
\label{sec:related}
CBR as a methodology is naturally interpretable due to its central role in harnessing episodic knowledge in the form of cases, where the assumption that cases of similar problems have similar solutions is innately easy for end users to understand and act upon. 
The link between XAI and CBR is not new, with some of the earliest works dating back to the eighties, where the focus was on explanation methods to support further interpretability in CBR i.e. XAI for CBR. 
Here good explanations were considered as those that took into account the explainer's goals and beliefs.

The need for explanations is specific to each user, and assisting them in expressing this need requires personalisation, considering both situational and individual elements, as highlighted by Leake~\cite{leake1988evaluating}.
Typically explanations were represented as part of the case structure using knowledge-rich rules and scripts to store explanation knowledge~\cite{schank1989creativity}. 
Explanation-based indexing commonly used features known to be good at explaining the cause of faults as indices for case retrieval~\cite{kass1986swale, barletta1988explanation, aamodt1993explanation}. 
In this way, indexing knowledge could also be used effectively to explain case retrieval and improve case ranking by combining causal knowledge.
In order to reduce the burden of gathering explanations at case creation, research began to focus on reusing past case-based explanations and manually adapting them to fit current anomalous situations needing explanation~\cite{leake1986organizing,schank2014inside}. 
This is sensible as similar problem situations are likely to benefit from similar explanations. 
This idea is similar to our iSee platform, which also exploits past explanation experiences for new situations. However, unlike these past works, explainers in iSee are aimed at explaining AI black box models, not CBR systems.

In recent work methods from CBR, specifically CBR's similarity knowledge, have been employed to generate factual, counterfactual, and semi-factual explanations for black-box models. 
Factual k-NN Explanations use model-specific neural prototypes~\cite{Li-aaai18} and model-agnostic twin systems ~\cite{kenny2019twin} to provide clear and understandable justifications for AI decisions.
In some scenarios, neighbours within the case base act as explainers ~\cite{doyle2004explanation}, providing locally faithful surrogates or twin explanations ~\cite{nugent2005case}.
Counterfactual k-NN Explanations also use similarity knowledge but to identify Nearest Unlike Neighbours (NUNs) for valid action recommendations helping users understand what minimal changes could alter an AI's decision~\cite{wiratunga2021discern, smyth2022few, brughmans2023nice}.
Semi-Factual k-NN Explanations combine factual and non-factual explanations (with Farthest Like Neighbours and NUN combinations) to offer a more comprehensive understanding of the decision-making process~\cite{vats2023changes,aryal2024semi}.

The adaptation step in CBR commonly includes constructive~\cite{plaza2002constructive} and transformative~\cite{craw2006learning} adaptation. Both of which we make use of within \isee to enable the reuse of past explanation experiences having tailored them to the current situation. Our work is unique in that we introduce adaptation operators applicable for explainer method reuse. 
The need for interactive XAI methodologies is closely linked to aligning with the evolving mental models of end users~\cite{hoffman2023measures}. iSee also employs a dialogue interface to facilitate interaction, guided by transitions prescribed by the explanation strategy. This ensures that explanation strategies are both effective and user-friendly.



\section{\isee Platform}
\label{sec:app}

The goal of the \isee platform is to help design users create and refine explanation strategies for their XAI systems to ensure end-user satisfaction.
Using the CBR 4R steps, \isee is organised to retrieve, reuse, revise, and retain explanation experiences as cases. 
Figure~\ref{fig:cbr} illustrates how the \isee platform is underpinned by the CBR paradigm. 
Central to CBR's 4Rs are its knowledge containers: case base, case similarity and case adaptation. In \isee, these containers are formalised using the \isee ontology for interoperability. 

\subsection{\isee platform overview}
The \isee Cockpit elicits explainability requirements from a design-user, who is an expert of the AI system's design and its stakeholder needs. These requirements form the query to our case base of past experiences, facilitating the retrieval of the most suitable explanation strategies. \isee provides tools to automatically adapt a recommended solution to further match design-user requirements by reusing multiple explanation strategies from nearest neighbours. 
The design user can then evaluate a recommended~(and adapted) strategy solution with a representative sample of their stakeholders to get feedback that can then be used for collaborative revision of the case description and solution. Once the stakeholder explanation needs are met, the design user can finalise the validated explanation strategy for their AI system thus forming a new case. 
The quality and coverage of cases in the case base enhances case-based recommendations. 
Accordingly retaining a complete anonymised case with the design user's consent is an important last step in \isee's 4R CBR cycle. 

The iSee platform was implemented using a micro-service based approach. Each module of the platform (i.e. user interface,  case retrieval, failure-driven adaptation, etc) can be hosted and executed independently on a single or multiple servers. Modules are logically connected to each other through standardised API endpoints, allowing flexibility for allocation of computational resources required to execute them.

\subsection{Explanation Experience Case Base}

\begin{figure}[t]
\centering
\includegraphics[width=.45\textwidth]{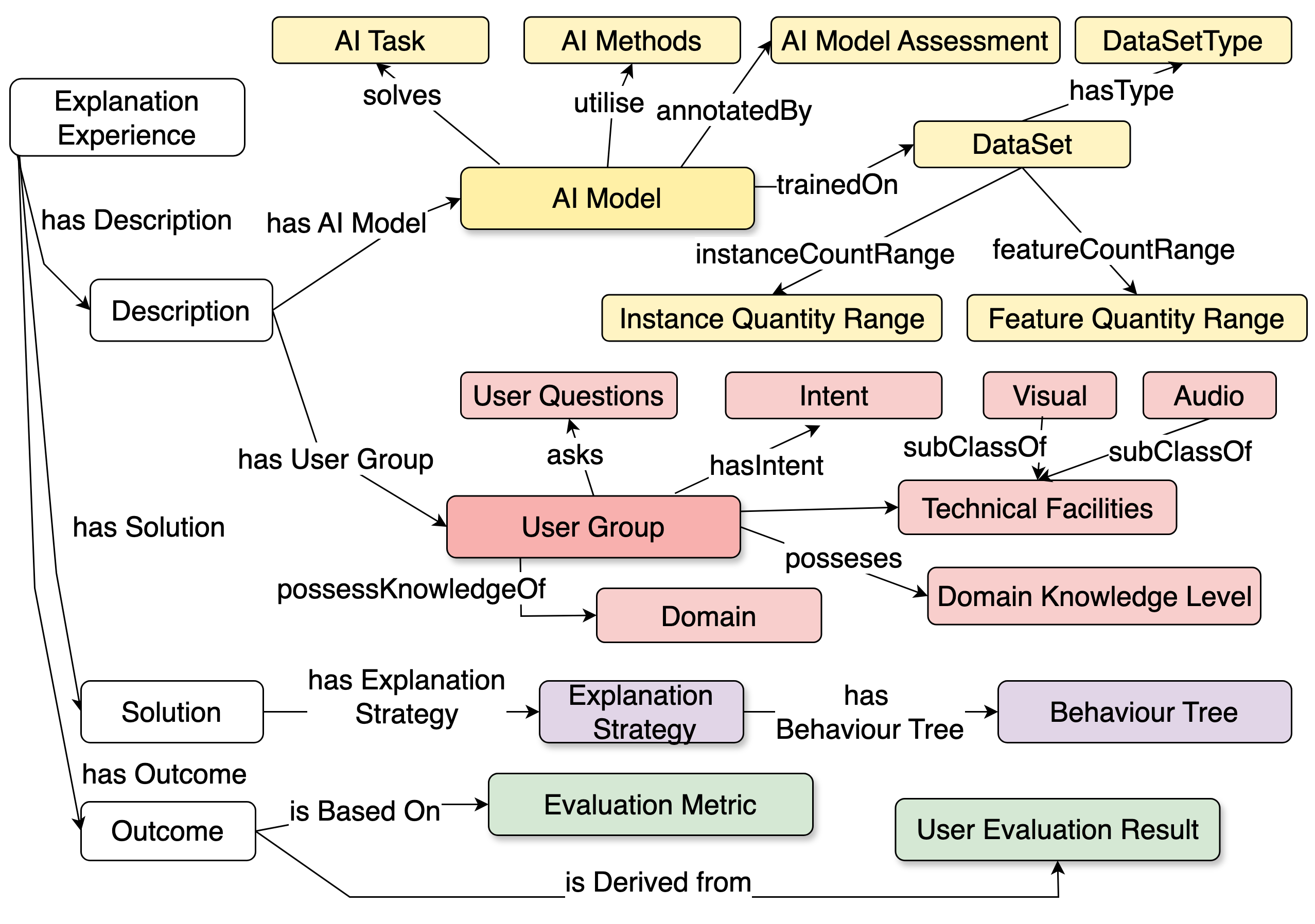}
\caption{Explanation Experience Case}
\label{fig:case}
\end{figure}

An explanation experience case is a multi-faceted entity that encapsulates several knowledge constructs: the attributes of the AI system; user groups and their explanation needs; the explanation strategy; and user explanation experience feedback~(see Figure~\ref{fig:case}). 
More formally
the iSee case base is a collection of past explanation experiences, each case $c$ represented as a triplet.
\[
    c = \{c_\mathcal{D}, c_\mathcal{S}, c_\mathcal{O}\}
\]
Where case description~($\mathcal{D}$) covers the constructs related to explanation requirements, a solution~($\mathcal{S}$) representing the explanation strategy and an outcome $\mathcal{O}$ capturing user feedback. 
Here a query $q$ is a case where the solution and the outcome are empty~($\mathcal{S},\mathcal{O}=\emptyset$). 
The majority of the cases are selected from literature following a critical review and we include several anonymised industry cases. \isee utilises these cases to recommend strategies to design users who are looking to build explainability in their AI systems. 

\begin{figure}[t]
    \centering
    \includegraphics[width=.45\textwidth]{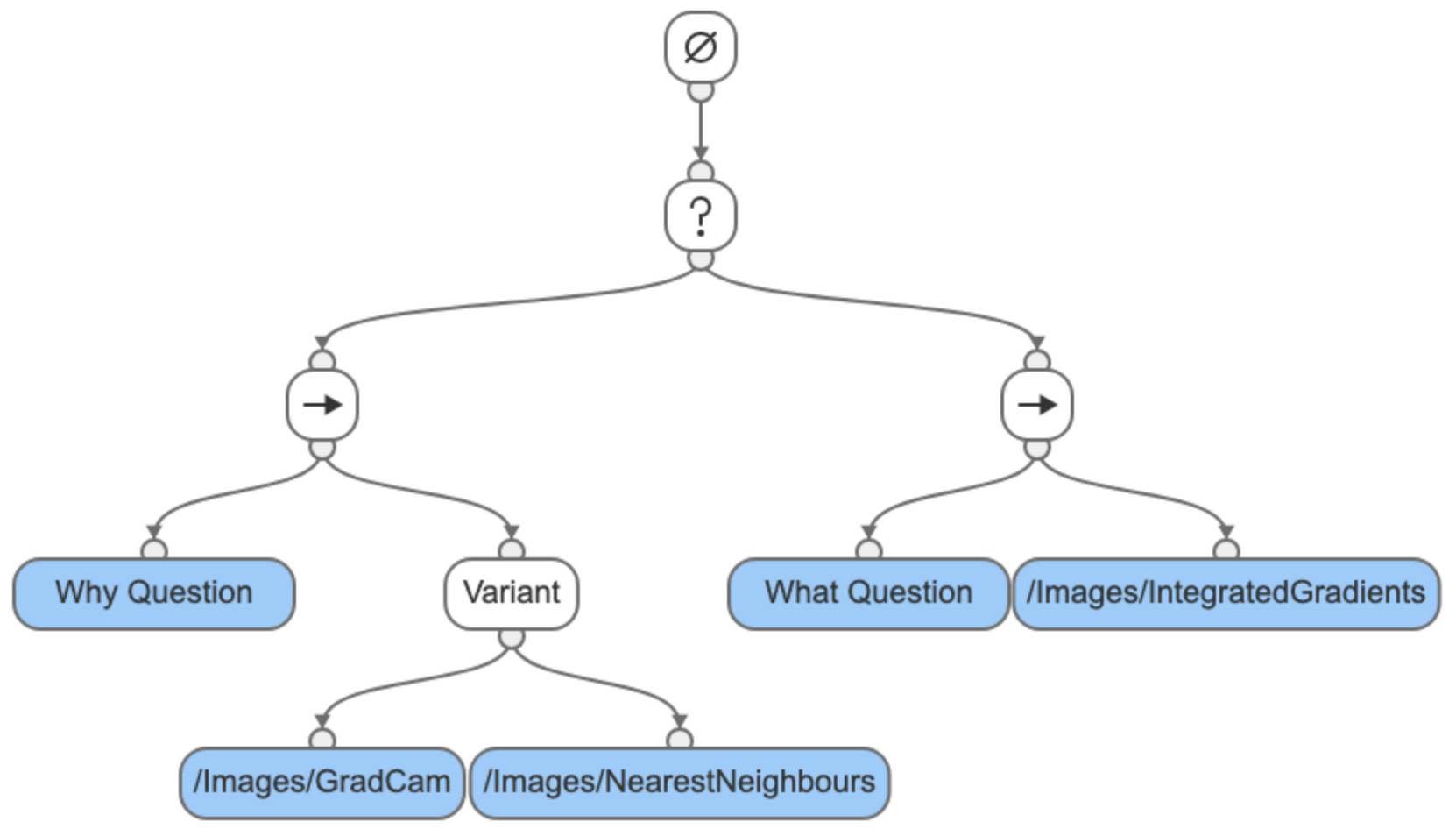}
    \caption{Sample Case Solution}
    \label{fig:sol}
\end{figure}

An explanation strategy, i.e, the case solution is modelled using a Behaviour Tree~(BT)~\cite{colledanchise2018behavior}. 
The example explanation strategy BT in Figure~\ref{fig:sol} is executed as follows. If the user asks a ``why'' kind of question, answer them with a GradCAM explanation and if they need to verify with another type of explanation~(variant) provide the nearest neighbours; if the user is still not satisfied and asks a further ``what'' kind of question, provide them an Integrated Gradients explanation. 
In this way, a BT model is a good way to manage execution in a controlled manner as it allows systematic handling of different types of user questions and corresponding explanation strategy sub-trees.

\subsection{Case Retrieval using Similarity Knowledge}
The retrieval task finds similar cases from the case base using similarity knowledge, which specifies local similarity metrics and aggregates them into a global similarity score. 
\[
    local\_sim = 
    \begin{cases}
    WP & \text{if } j \in [\textit{AI Task}, \textit{AI Method}]\\
    QI & \text{if } j \in [\textit{Technical Facilities}, \textit{User Questions}]\\
    EM & \text{otherwise}
\end{cases}
\]
where $j$ refers to a feature in $\mathcal{D}$. The similarity metrics are as follows.
\begin{description}
\item [Wu \& Palmer~(WP)] is a taxonomy path-based similarity metric originally implemented for calculating word similarities. For \emph{AI Task} and \emph{AI Method} case attributes, given the taxonomic representation from \isee ontology, it computes node similarity between the query and the case nodes based on node depths and distances from the most specific common ancestor~\cite{nkisi2022adapting}.
\item [Query Intersection~(QI)] is applicable for attributes where the data type is a set of ontology individuals such as attributes \emph{User Questions} and \emph{Technical Facilities}. 
Given the feature $j$ from query $q$ and case $i$, it calculates the similarity as the intersection between two sets normalised by the query set size as $(|c^q_j \cap c^i_j|)/|c^q_j|$.
\item [Exact Match~(EM)] similarity indicates a string match. This is applied both for case attributes that are ontology individuals, and is the most common method of comparison. 
\end{description}

\isee implements retrieval using CloodCBR~\cite{nkisi2022adapting} in two phases: 1) filter case base to only include cases that exactly match the query \emph{DataSetType}~($dt$)~(Equation~\ref{eq:filter}); and 2) calculate pair-wise similarity to each filtered case to select the top k most similar cases~(Equation~\ref{eq:agg}). 



\begin{equation}
    \mathcal{C}' = \{c^i \in \mathcal{C} \text{  }| \text{  }c^q_{dt} = c^i_{dt}\}
\label{eq:filter}
\end{equation}

\begin{equation}
     \text{Top-}k = \underset{c^i \in \mathcal{C}'}{\operatorname{argmax}}^k \left\{\frac{1}{|\mathcal{D}|} \sum_{j=1}^{|\mathcal{D}|} local\_sim(c^q_j, c^i_j)  \right\}
\label{eq:agg}
\end{equation}
The similarity between the query case $c^q$, and a case $c^i$ from the case subset $\mathcal{C}'$ is calculated as the aggregation of local similarities as in Equation~\ref{eq:agg}. 
The single top case, is the case with the highest global similarity score among the top \( k \) and the recommended explanation strategy for the query requirements. 

\subsection{Explanation Strategy Reuse}
The CBR methodology recommends solution adaptation before reuse to: 1) address unmet requirements on the query description; and 2) personalise the solution utilising domain knowledge. \isee offers a failure-driven adaptation algorithm to address the former and envision the latter to be a manual process. 

\begin{figure}[t]
    \centering
    \includegraphics[width=.5\textwidth]{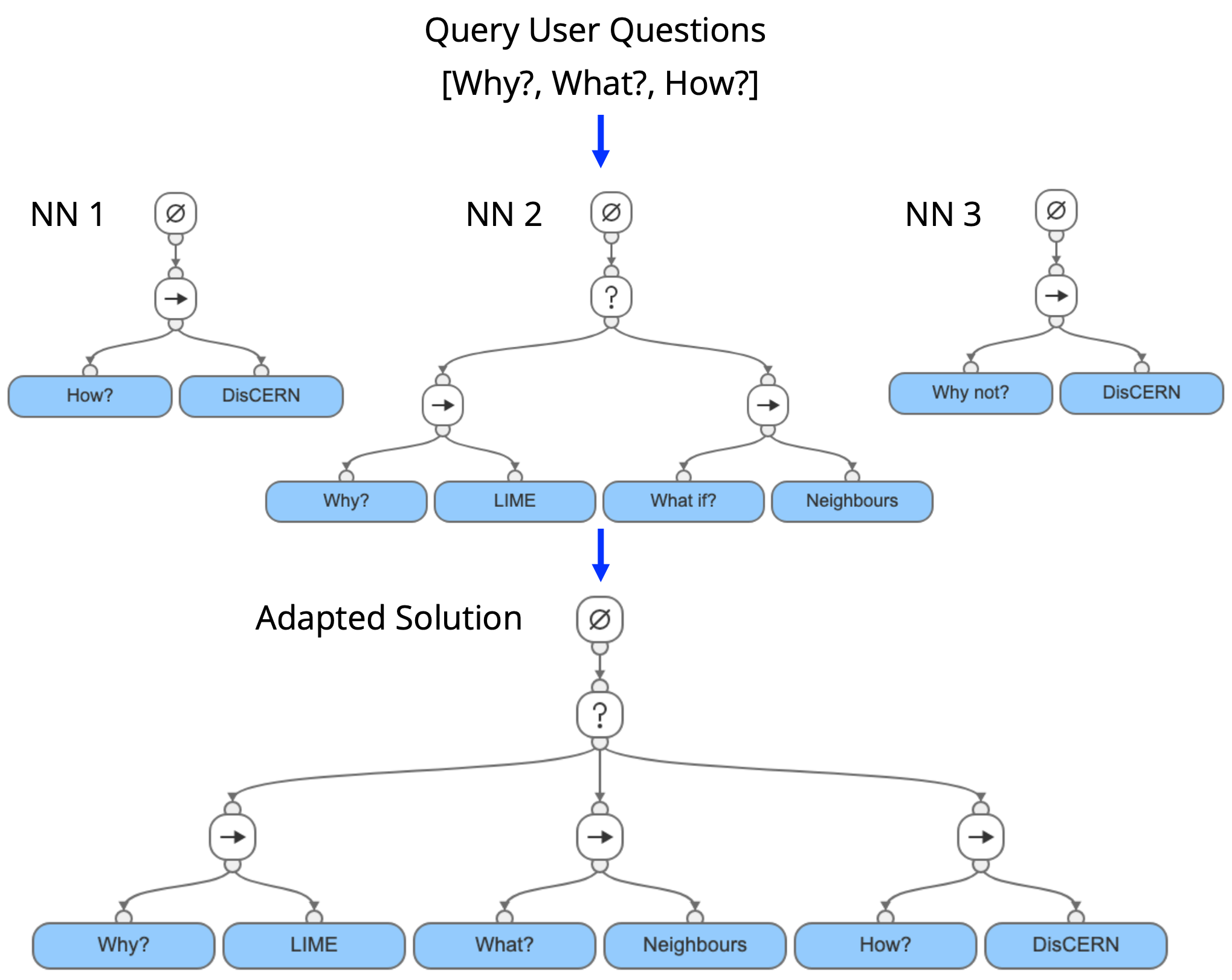}
    \caption{Failure Driven Adaptation}
    \label{fig:fda}
\end{figure}

Adaptation of solutions is driven by the failure to fulfil the query's \emph{User Questions}. The mismatch between the recommended case and the query is calculated using the Query Intersection similarity and when similarity is $ \le 1$ we apply a stable marriage algorithm on the top k case solutions~$(k > 1)$ to find user question-explanation strategy sub-trees that satisfy all (or as many) of the user questions that appear in the user's query. The \textit{adapted} explanation strategy BT is formed of these sub-trees. 
Figure~\ref{fig:fda} presents an example where 2 of the query user questions ("Why" and "What") are not met by the recommended case solution~(NN1). Accordingly, \isee uses the top 3 neighbours to find best sub-tree matches in neighbours 2 and 3 to form the adapted solution~\cite{nkisi2023failure}.

\subsection{Explanation Strategy Revision}
\isee provides an editor and the following supporting tools for design users to revise explanation strategies. 

\begin{description}
\item [Explainer Applicability ]warns the design user about the implementation mismatches between the explainers in the recommended strategy and their query case. These mismatches include 1) the implementation framework supported by the explainers and that of the query AI model; 2) model access requirements of the explainers and model access provided by the design user~(model file or API access to the predict function); and 3) labelled data requirements of the explainers and data provided by the design user. 

\begin{figure}[t]
    \centering
    \includegraphics[width=.45\textwidth]{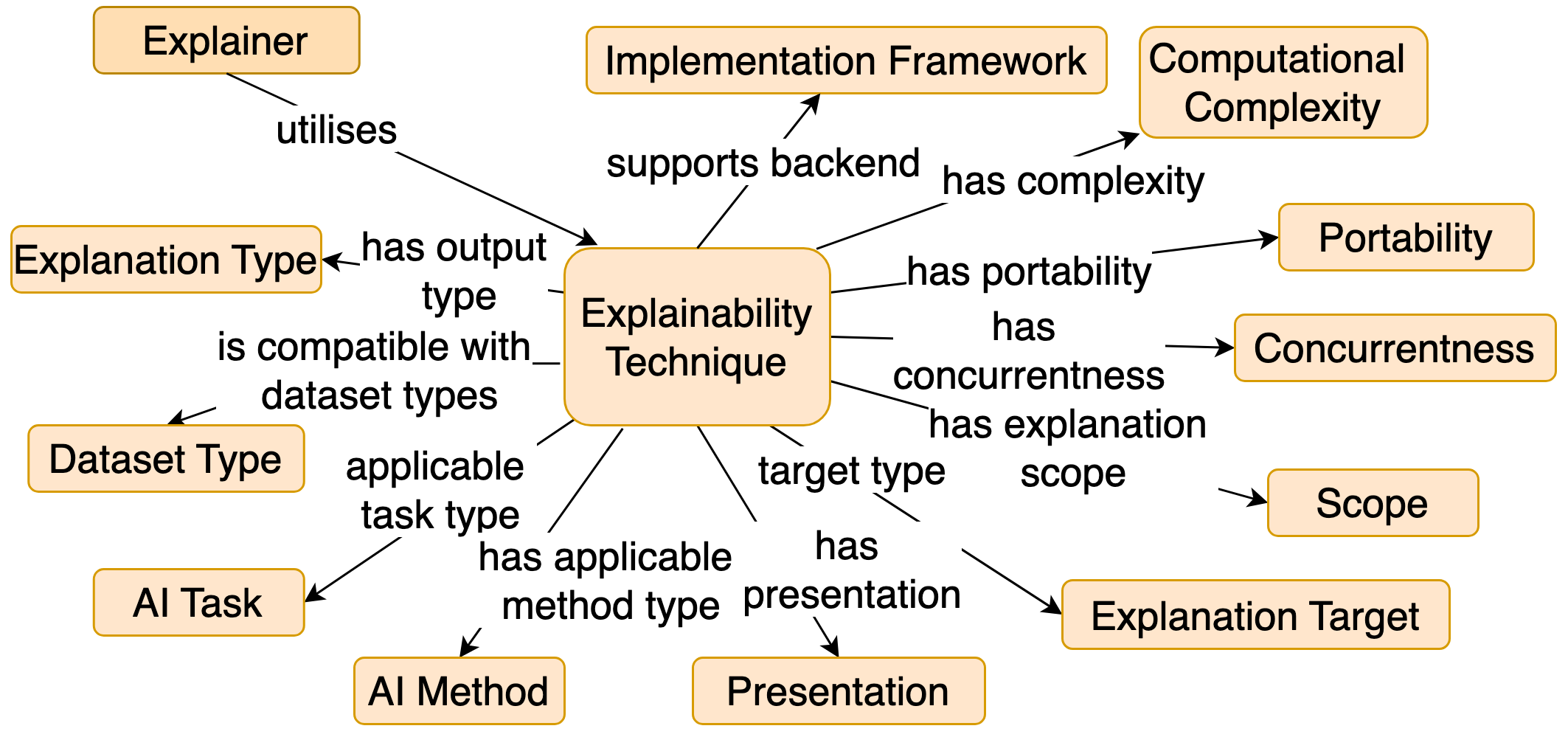}
    \caption{Explainer Properties}
    \label{fig:ex}
\end{figure}

\item [Explainer Substitution ] provides the design user with substitution recommendations for a selected explainer. This follows a similar approach to retrieval on a library of explainers where each is characterised using a set of semantic features~(see Figure~\ref{fig:ex}). The similarity between explainers is calculated as $e\_sim(e^q,e^i) = \frac{1}{|\mathcal{M}|} \sum_{j=1}^{|\mathcal{M}|} local\_e\_sim(e^q_j, e^i_j)$ using the following local similarities.
\[
local\_e\_sim = 
    \begin{cases}
    WP & \text{if } j \in [\textit{AI Tasks}, \textit{AI Methods}, \textit{Presentation}]\\
    QI & \text{if } j \in [\textit{Implementation Frameworks}, \\
    &  \textit{Explanation Technique}, \textit{Explanation Type}]\\
    EM & \text{otherwise}
    \end{cases}
\]
\item[Sub-tree Substitution ] provides applicable sub-tree substitutions for a selected sub-tree. Substitutions are selected from the solutions of similar cases based on edit-distance similarity. \isee transforms the query and case sub-trees into directed graphs and calculates edit distance using the node similarities defined below where $type(n)$ returns the strategy node type.
\[
    node\_sim = 
    \begin{cases}
    e\_sim & \text{if } type(\{n^q, n^i\}) = \text{Explainer}\\
    sem\_sim(.) & \text{if } type(\{n^q, n^i\}) = \text{User Question}\\
    1 & type(n^q) = type(n^i)\\
    0 & type(n^q) \neq type(n^i)\\
\end{cases}
\]

\end{description}

\subsection{Case Retention}
Once the design user has a recommended solution, adapted and/or revised, they evaluate it with their stakeholders. \isee provides a chatbot interface where stakeholders can execute the strategy to create explanation experiences and provide feedback~\cite{wijekoon2022behaviour,wijekoon2024tell}. Design users can utilise this feedback to iteratively improve the case description and the solution, aiming for stakeholder satisfaction. 

We form the case outcome from the stakeholder feedback obtained using the XAI Experience Quality (XEQ) Scale~\cite{xeq} tool which was developed as a psychometric scale for measuring the quality of explanation experiences across 4 dimensions: Learning, Utility, Fulfilment and Engagement. The case outcome records the mean score in each dimension.
During case retention, \isee creates an anonymised copy of the complete case and retains it in the case base. A case maintenance policy~\cite{craw2007informed} can then be used to periodically review the case base considering case coverage. 


\begin{figure*}
    \centering
    \includegraphics[width=0.98\textwidth]{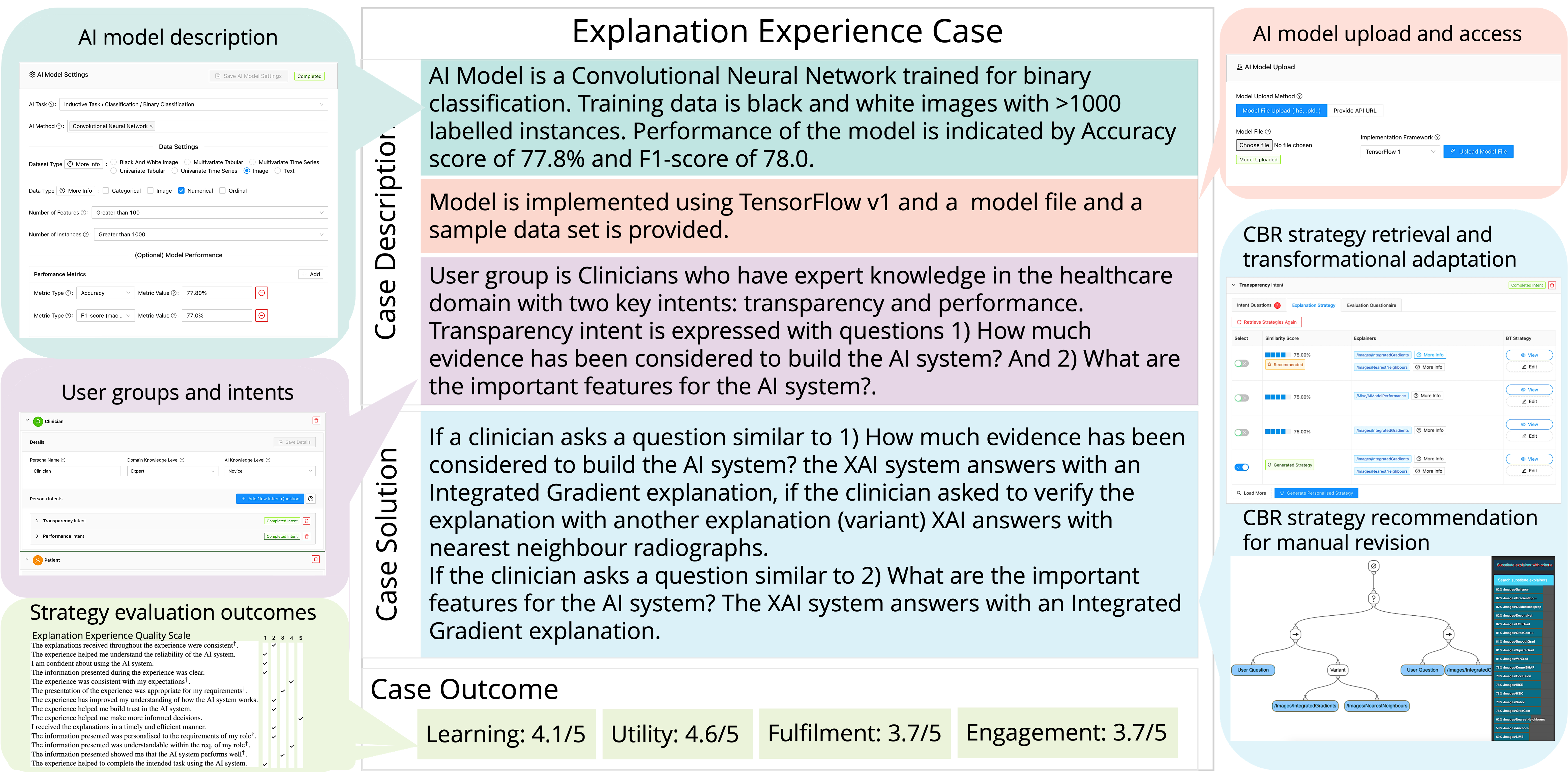}
    \caption{Radiograph Classification Use Case created following the \isee CBR 4Rs - Please refer to the supplementary material to access the full-resolution screenshots of the platform}
    \label{fig:jiva}
\end{figure*}

\begin{table*}[ht]
\centering
\caption{Design User Profiles}
\label{tbl:design-profiles}
\footnotesize
\renewcommand{\arraystretch}{1.2}
\begin{tabular}{p{2cm}p{9cm}p{2cm}p{3cm}}
\hline
ID & Participant Description & Experience in AI & Experience in XAI\\
\hline
D1&Academic Researcher, designed and developed the AI system&Proficient&Novice\\
D2&Lead for AI system delivery within the company, involved in the requirements management and design of the AI system&Expert&Novice\\
D3&Lead data scientist of the company involved in the requirements management, design and development of the AI system&Expert&Expert\\
D4&Academic Researcher, designed and led the development of the AI system&Expert&Proficient\\
D5&Academic Researcher, designed and led the development of the AI system&Expert&Expert\\
D6&Lead data scientist of the company involved in the requirements management, design and development of the AI system&Expert&Expert\\
\hline
\end{tabular}
\end{table*}

\subsection{Example Use Case: Radiograph Classification}
We describe a radiograph classification system provided by an industry partner. The AI system is implemented using ConvNet-based architecture for binary classification of fractures in radiographs. The stakeholder explanation needs of this use case stem from two factors: 1) to improve the quality of their product for end-users; and 2) to increase regulatory compliance with relevant governance bodies. The design user described two user groups: 1) clinicians who are using the AI system for decision support; and 2) managers who are looking to evaluate the compliance, risk and regulatory requirements. 

Using callouts of \isee screenshots in Figure~\ref{fig:jiva}, we illustrate how a design user can interact with the \isee \textit{retrieve}, \textit{reuse}, and \textit{revise} tools to create a complete Explanation Experience case, containing both the case description and solution parts, and \textit{retain} it in the case base.
Firstly, an AI model description and implementation of a ConvNet model for binary classification of black and white radiography images is entered into the \isee Cockpit. Further details on how to access the model can also be provided.
User groups and intents part of the description include 
details of a clinician persona, alongside corresponding intents in transparency and performance, thus completing the query case description $q_\mathcal{D}$.
The completed case description parts can be used to query the \isee case base and \textit{retrieve} a set of candidate cases containing previous best practices of explanation strategies.
In the example in Figure~\ref{fig:jiva}, of the retrieved three cases containing variations of strategies include a combination of feature attribution and nearest neighbour-based explainers (top of blue callout). 
The design user can decide to \textit{reuse} the recommended solution arrived at after \isee performs a failure-driven transformational adaptation to obtain a personalised strategy. 
After that, they can decide to perform a manual \textit{revise} step using a strategy editor (bottom of blue callout), which will provide a ranked list of substitute explainers for any selected explainer node that the designer user wishes to change~(as demonstrated here by highlighting the Integrated Gradients explainer node for substitution). 
Once revisions are complete, the case contains the refined solution component. It can be evaluated with target stakeholders to identify the case outcome~(which is measured against the dimensions of the XEQ Scale). This allows the formation of a complete case, which can subsequently be \textit{retained} in the case base to inform future practice.    

\section{User Experience Evaluation with Design Users}
\label{sec:design}
A summative assessment, using a mixed methods study, was performed to evaluate the user experience of design users. We aim to evaluate the following two dimensions. 
\begin{itemize}
\item Utility: Do the design users perceive the tool as fit for purpose?
\item Usability: Do the design users find it easy and efficient to complete their tasks?
\end{itemize}

\subsection{Study Protocol}
We planned a two-stage user-centred evaluation session with a design user lasting approximately 1 hour. A session is standardised using the following protocol:
\begin{enumerate}
\item To open the session, the researcher provided a brief overview of the \isee project and the objectives of the session. A toy example of a loan approval XAI system was used to illustrate the specific information requested on the Cockpit.
\item We then conducted a concurrent Think-Aloud Protocol (TAP) where the participant was given access to the \isee Cockpit and asked to create their use case as a design-user of the system. Throughout the session, participants were encouraged to vocalise their thoughts. The researcher intervened only when necessary, (i.e. when the user sought clarification or was unable to proceed). 
\item On completion of the TAP, the participant was asked to respond to the User Experience Questionnaire~(UEQ) consisting of 26 questions on a 7-step Likert scale. 
\item To conclude the session, the researcher asked participants a series of open-ended questions. These questions aimed to establish a design user profile and capture any additional comments or insights regarding their experience.
\end{enumerate}

\begin{figure}[t]
\centering
\includegraphics[width=.5\textwidth]{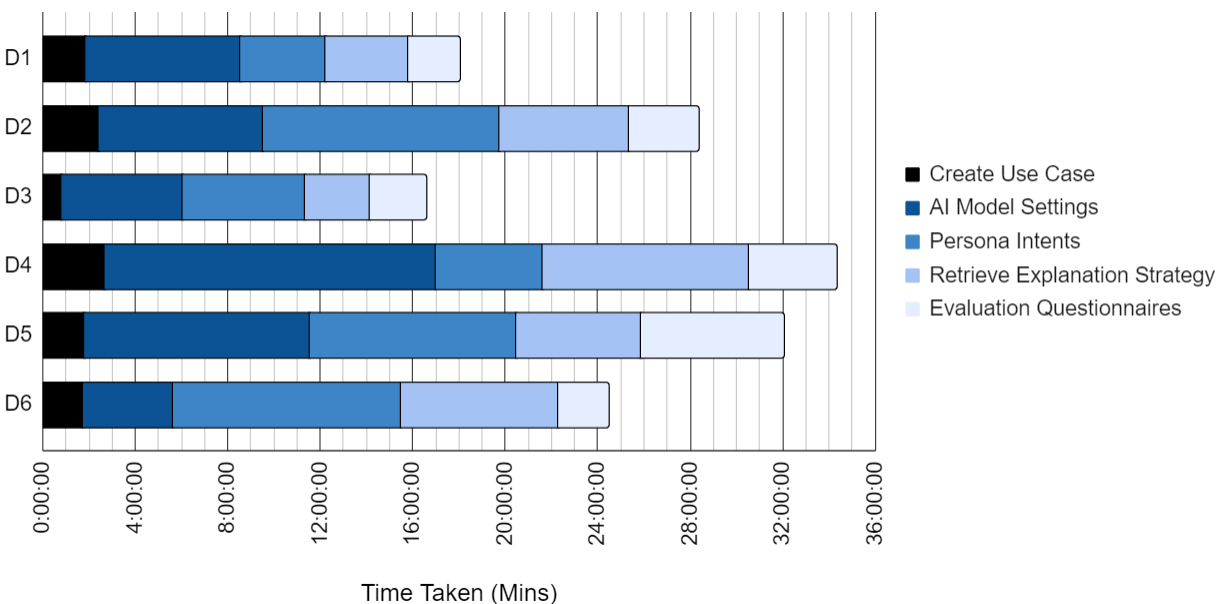}
\caption{\isee Cockpit time taken per component}
\label{fig:time-taken}
\end{figure}
\subsection{Recruitment}
This study involved six design users, two conducted in person and four online over Microsoft Teams. They were recruited through existing academic and industry connections with the leading institution. These design users were distinct from any design users who were involved in the initial UI/UX design activities to avoid bias. At the start of the study, the leading researcher obtained informed consent from the design user to record the screen and audio during the session and to use the data generated during the session exclusively for research purposes. 
The design user profiles are summarised in Table~\ref{tbl:design-profiles}. In the rest of the paper, the ID column is used to reference each design user.

\subsection{Study Instruments} 
The UEQ questions assessed user perception of usability~(UEQ dimensions efficiency, perspicuity and dependability) as well as user experience~(UEQ dimensions novelty and stimulation). The overall impression of the product is measured by the attractiveness dimension. For our study, we were interested in 2 out of the 4 objectives of the original UEQ framework related to establishing sufficient user experience and identifying areas of improvement. For a more detailed discussion on the UEQ, please refer to~\cite{schrepp2014applying}.

The UEQ benchmark is based on an analysis of 163 products consisting of business applications, development tools, e-commerce websites, social networks and mobile applications. Each dimension measures the scores in 5 categories from \textit{excellent}, \textit{good}, \textit{above average}, \textit{below average} and \textit{bad}, and a new product is expected to reach \textit{good} category in all dimensions. We note that the benchmark has not considered technological or research-oriented products like the \isee platform. Also, due to the limited number of use cases, we are not able to establish the statistical significance of the results.
The Concurrent TAP method was used to obtain qualitative insights into user experience by asking users to verbalise their thoughts as they use a system to perform a specific task~\cite{jaaskelainen2010think,van2003retrospective}. These sessions facilitate targeted usability evaluation of specialist tools, as they allow flexible task performance and researcher intervention when required.

\subsection{Analysis}
Using the above instruments, the study generated three artefacts: 1) transcriptions of the audio and screen recordings; 2) researcher notes documented during TAP sessions; and 3) UEQ responses. They are utilised in a two-part quantitative and qualitative analysis: 1) measure user experience against established UEQ benchmarks; and 2) combine UEQ responses, the researcher notes, and transcripts to perform a thematic analysis of TAP session outcomes. 

We used the recording to analyse the time taken to use the cockpit to produce a functional explanation strategy. The starting point was when participants clicked the \textit{'Create Use Case'} button, and the endpoint was when participants saved the evaluation questionnaire (which is the final stage of use case creation). The mean time taken was 25 minutes and 51 seconds and a breakdown of time taken per component of the Cockpit is available in Figure~\ref{fig:time-taken}.

Despite the relative freedom of the TAP session, all users converged on a similar progression through the components of the Cockpit interface. The similarity of user pathways highlights the interface is structured in a logical manner. 
Examining individual sessions, all 3 industry design users spent the majority of their time~(\%) identifying persona intents while academic design users primarily focused on configuring AI model settings. This suggests a difference in prioritisation for explanation strategy design where industry users are focused on end-user needs, while academics are focused on model details. We highlight that differences in expertise level do not seem to reflect in the time taken to complete the exercise. This promising result highlights the platform is equally suitable for expert and non-expert design users.  
\section{Results and Discussion}
\label{sec:results}

\subsection{UEQ Findings}
\begin{figure}[t]
\centering
\includegraphics[width=.5\textwidth]{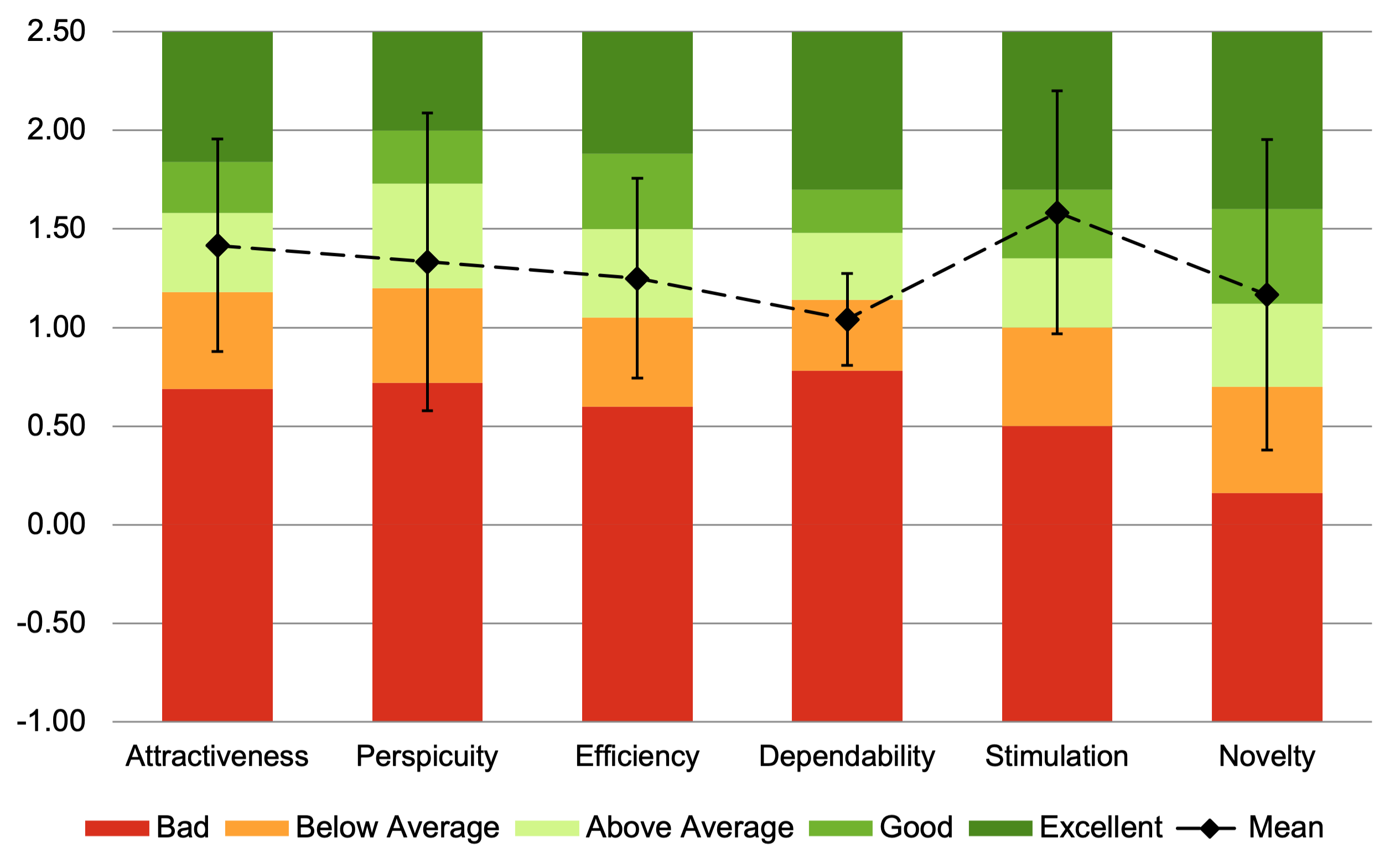}
\caption{\isee Cockpit user experience against UEQ Benchmark}
\label{fig:design-benchmark}
\end{figure}

Overall, the Cockpit has been scored above \textit{above average}, and achieves \textit{good} category on user experience (i.e. Stimulation and Novelty). Figure~\ref{fig:design-benchmark} presents how the \isee Cockpit scored across the six dimensions measured by UEQ. The y-axis scale ranges from +3 to -3 with the mean response of 0 representing a \textit{neutral} sentiment. For Attractiveness, Perspicuity and Efficiency, the cockpit scores \textit{above average}, while scoring \textit{below average} for Dependability.  
It is noteworthy that the benchmark has been established on products intended for the general public whereas the Cockpit caters to a specialised group of expert users. Despite this, we have achieved \textit{above average}, which is very promising.

Figure~\ref{fig:answers} presents a detailed view of individual responses. We highlight that responses are overall positive; 20/26 questions are \textit{majority positive}, with the remaining 6 questions being \textit{majority neutral} or equally \textit{split} between positive and neutral. Importantly, there are no questions with majority-negative responses~(with only negative or neutral responses). We explore the justification for these responses by examining the TAP session transcripts in the following section.

\subsection{TAP Findings}
Here we present the findings of a thematic analysis broken down across the UEQ themes and evidenced using TAP session transcripts. 

\begin{figure}[t]
\centering
\includegraphics[width=.5\textwidth]{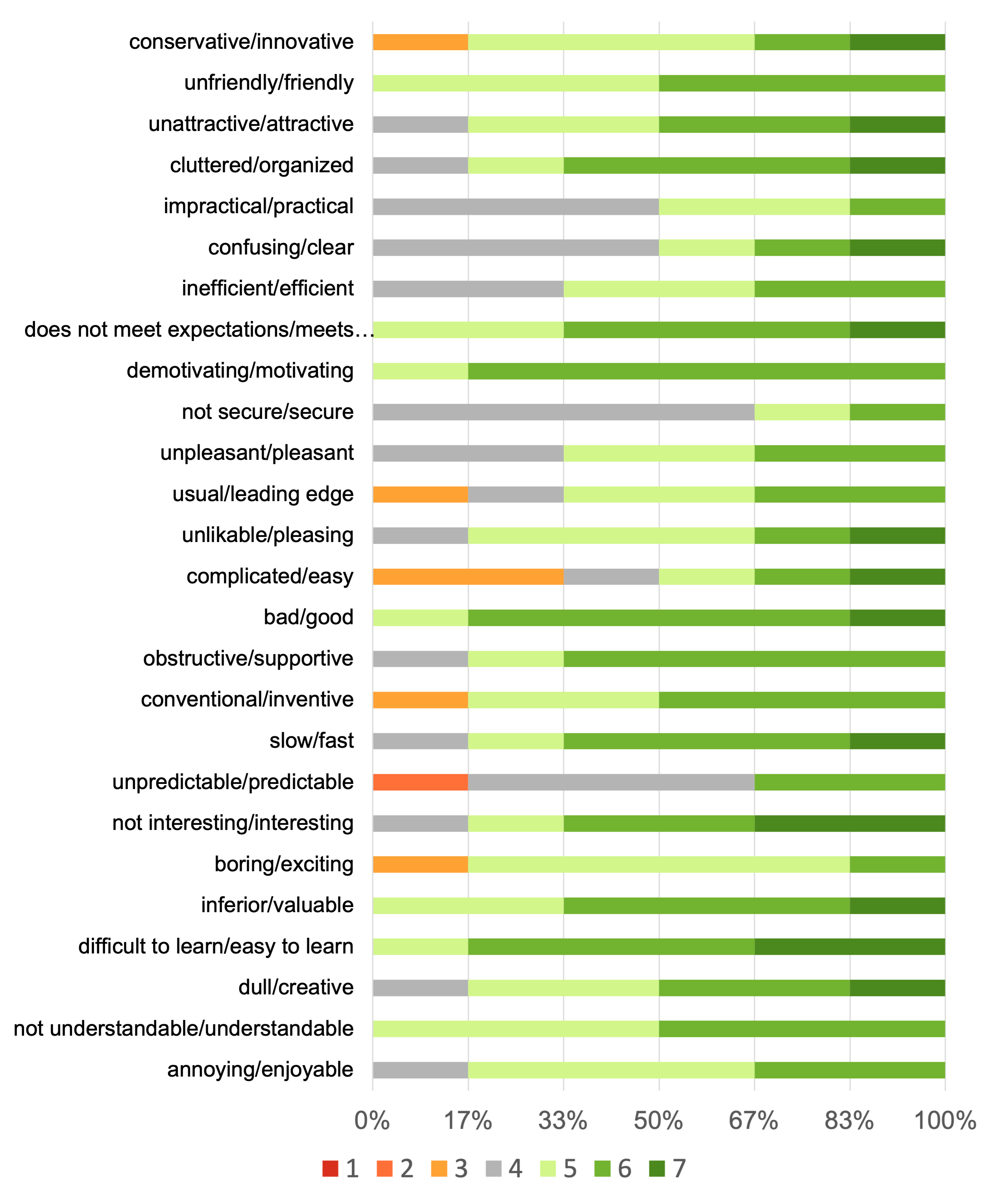}
\caption{UEQ individual response distributions}
\label{fig:answers}
\end{figure}
\medskip
\noindent \textbf{Attractiveness}
Users had an overall positive view of the \isee Cockpit, and quickly identified the value that stakeholder-driven explanation strategies would add to their practice:
\begin{quote}
\small
    \textit{Each time [an AI] gives a prediction, I don’t trust it at face value. So getting the the evidence behind the prediction is one of the high priorities for me.} - D3
\end{quote}
Similar comments were made by all users. We take this as evidence the \isee Cockpit contributes to satisfying business needs for low-code development of explanation strategies. 
There were comments regarding the presentation of the Cockpit interface. Most comments targeted improving guidance for navigation or reducing verbiage~(see below discussion of Perspicuity for examples). Otherwise, users were satisfied with the presentation. 

\medskip
\noindent \textbf{Perspicuity}
All users required prompting at different stages of using the \isee system. The Cockpit contains a large volume of information, which users sometimes find complicated or intimidating (as evidenced in Figure~\ref{fig:answers} responses for complicated/easy). This was evidenced throughout the TAP sessions (\textit{``These are not common words."} - D1; \textit{``There is a lot here"} - D3). Despite this, we highlight that all users responded very positively to the UEQ statement the system is easy to learn. This highlights that users feel the Cockpit will be easier to use in subsequent iterations, but requires more scaffolding on first use. To resolve this, we plan to develop a set of tutorials to improve the initial experience of using the system.   

\medskip
\noindent \textbf{Efficiency}
Users found that using the Cockpit was efficient. D5 observed that pre-filled components derived from the ontology facilitated data entry:
\begin{quote}
\small
    \textit{I really like that you have these options to kind of pick from. I think it makes that much easier.} - D5
\end{quote}
The information from the ontology therefore provides both standardisation of data input for case formalisation, as well as user support. 
From observing the video recordings of TAP sessions, we identified that navigation of the interface required direction. This mostly impacted areas where the subsequent task was not clear (for example, several users did not realise that an explanation strategy was retrieved for each persona intent). We will address this with added navigation support in future versions of the system.

\medskip
\noindent \textbf{Dependability}
The \isee cockpit scored \textit{Below Average} on the Dependability dimension, and one reason for this was security was highlighted as a key concern (highlighted by the \textit{majority-neutral} response to not secure/secure in Figure~\ref{fig:answers}). Three use cases in this study were developed as open-science and the others were AI systems developed for proprietary use. Design users from both categories raised questions about security and privacy. For example, D2 highlighted the need for authenticating access to model APIs,
and D1 requested confirmation of web page secuity (\textit{``is it HTTPS?"}) before providing any input. 
Despite this, design users expressed satisfaction over features such as encrypted network communication and authentication provided by the Cockpit. 

\medskip
\noindent \textbf{Stimulation}
As evidenced in Figure~\ref{fig:answers}, all users found the Cockpit highly motivating. D4 highlighted that the capability of \isee  to facilitate comparison of multiple explainers would be very useful for testing, validating and understanding the model:
\begin{quote}
\small
    \textit{If I had a tool like this, it would be really helpful. When you develop [explanations], sometimes you have this kind of error analysis and if you have this kind of tool where you have several explanations, it's really, really interesting. Sometimes I had only one explainer and sometimes it doesn't work or it is biased, or focuses on things that are not really relevant.} - D4
\end{quote}
Similarly, D1 and D2 highlighted that the interaction had motivated them to test the tool with actual users:
\begin{quote}
\small
    \textit{Yeah, that's fantastic. I have a really nice explanation. As an AI developer I can create my personas and such that I can then get feedback on the particular explainers.} - D2
\end{quote}
We highlight these factors as evidence of the \isee system's utility, as inspiring users to develop and refine their explanation strategies is a key feature of the tool.

\medskip
\noindent \textbf{Novelty}
Feedback from users emphasised they found the novel aspects engaging. Specifically, users highlighted that the wealth of different explainers was exciting: 
\begin{quote}
\small
    \textit{You have a lot of techniques there, which I'm going to have to go and have a look at.} - D2
\end{quote}
Additionally, users expressed an interest in the underpinning methodology. D6 engaged in a discussion regarding CBR and the opportunities for empowering different explanations at an instance level (i.e. explaining complex instances using different explainer algorithms from simple instances). This evidences that the \isee platform encourages creative thinking and supports the sharing of best practices.

Overall, we found the results of the TAP sessions to be positive with encouragement for improvements. User comments actively supported the UEQ responses and highlighted user experience of the \isee Cockpit was equally satisfying for expert and non-expert users across use cases. Improving the usability of the interface shall be a key target in the ongoing development. Outcomes of the evaluation emphasise the need for clearer navigation support to facilitate interaction within the Cockpit. Finally, the perceived security of the interface will be improved, scoring better on the Dependability dimension. We will address these as part of ongoing development in the \isee system.  

\section{Conclusion}
\label{sec:conc}
In this paper, we have described the implementation and evaluation of the \isee system. The \isee platform is based on CBR for the reuse of best practices in creating multi-shot explanation experiences. We presented the findings of a comprehensive user-centred evaluation including both industry and academic participants where we demonstrated the utility and usability of the system via the recognised UEQ benchmark, and analysis of think-aloud session outcomes. Our findings highlighted that both expert and non-expert design users found \isee comparably useful in assisting the implementation of multi-shot XAI in their AI systems. In future, the \isee platform will continue to grow by evaluating and improving the \isee Cockpit to enhance the design user experience; improving the coverage of the case base for improved recommendations and extending the availability of explanation methods for improved adaptation and revision.  

\section*{Ethical Statement}
The study protocol was reviewed and approved by the lead institution's ethics review committee. Informed consent was obtained from all participants. 

\begin{ack}
The authors would like to thank Jiva.ai for providing the radiograph classification use case described in this paper, and all design users who participated in the user experience evaluation. 

iSee is an EU CHIST-ERA project which received funding for the UK from EPSRC under grant number EP/V061755/1; for Ireland from the Irish Research Council under grant number CHIST-ERA-2019-iSee; for France from The French National Research Agency under grant number ANR-21-CHR4-0004 and for Spain from the MCIN/AEI and European Union ``NextGenerationEU/PRTR'' under grant number PCI2020-120720-2.
\end{ack}

\bibliography{ref}

\section*{Supplementary Material}

\begin{figure*}
    \centering
    \includegraphics[width=\textwidth]{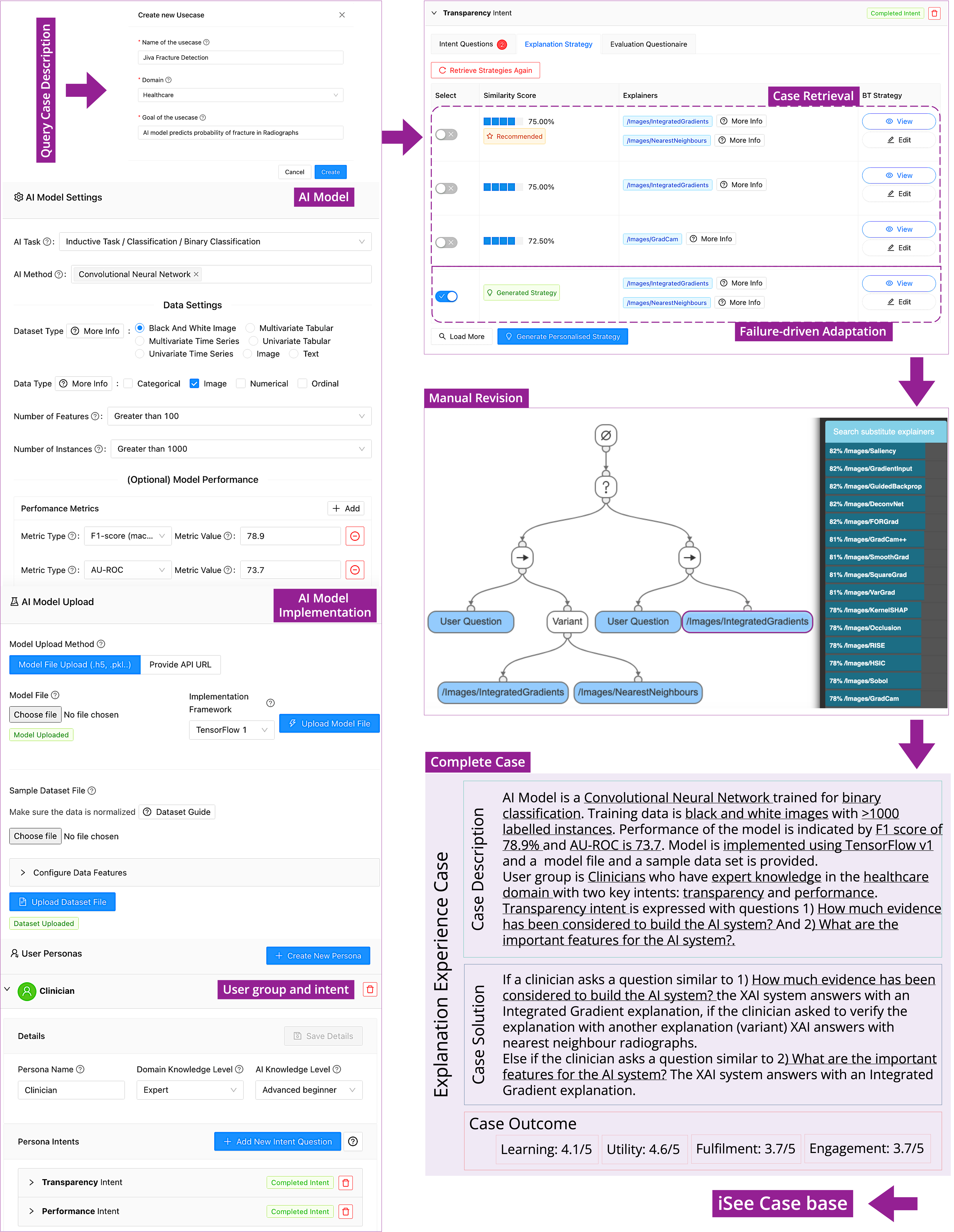}
    \caption{Full resolution platform screenshots - creating the Explanation Experience use case}
    \label{fig:screens}
\end{figure*}

The iSee platform can be accessed at \textit{https://cockpit.isee4xai.com/}. To request login details, please contact \texttt{hello@isee4xai.com} and follow the tutorials available at \textit{https://isee4xai.com/}. The complete platform code is publicly available on GitHub at \textit{https://github.com/isee4xai}. Figure~\ref{fig:screens} depicts how the platform facilitates the creation of an explanation experience use case. 

\end{document}